\documentclass[letterpaper, 10 pt, conference]{ieeeconf}  

\IEEEoverridecommandlockouts                              

\newcommand{\RNum}[1]{\uppercase\expandafter{\romannumeral #1\relax}}
\usepackage{graphicx,wrapfig,fullpage,amsmath,hhline,epsfig,verbatim,url,amssymb,multicol,multirow,cite}
\usepackage{times,color,soul} 
\usepackage[left=0.75in,top=0.70in,right=0.75in,bottom=0.80in]{geometry}
\usepackage{amsbsy,latexsym,mathrsfs,mathtools}

\usepackage{mathptmx} 
\DeclareMathAlphabet{\mathcal}{OMS}{cmsy}{m}{n}
\usepackage{bm}
\usepackage{float}
\usepackage{color,soul}
\usepackage{dsfont}
\usepackage{comment}
\usepackage{algorithm}
\usepackage{algorithmic}
\usepackage{psfrag}   

\usepackage{url}
\usepackage{setspace}
\usepackage[table]{xcolor}
\usepackage{paralist}
\usepackage{booktabs}
\usepackage{balance}

\title{Extended Capture Point and Optimization-based Control for Quadrupedal Robot Walking on Dynamic Rigid Surfaces} 

\author{Amir Iqbal and Yan Gu
	\thanks{Amir Iqbal and Yan Gu are with the Department of Mechanical Engineering, University
of Massachusetts Lowell, Lowell, MA 01854, U.S.A. Emails: {amir\_iqbal}@student.uml.edu and {yan\_gu}@uml.edu.
Research by A. Iqbal and Y. Gu was supported in part by NSF under Grant no. CMMI-1934280. Corresponding author: Y. Gu.
}
}

\begin{document}
	
	\maketitle
	\thispagestyle{empty}
	\pagestyle{empty}

\begin{abstract}         
Stabilizing legged robot locomotion on a dynamic rigid surface (DRS) (i.e., rigid surface that moves in the inertial frame) is a complex planning and control problem.
The complexity arises due to the hybrid nonlinear walking dynamics subject to explicitly time-varying holonomic constraints caused by the surface movement.
The first main contribution of this study is the extension of the capture point from walking on a static surface to locomotion on a DRS as well as the use of the resulting capture point for online motion planning.
The second main contribution is a quadratic-programming (QP) based feedback controller design that explicitly considers the DRS movement.
The stability and robustness of the proposed control approach are validated through simulations of a quadrupedal robot walking on a DRS with a rocking motion.
The simulation results also demonstrate the improved walking performance compared with our previous approach based on offline planning and input-output linearizing control that does not explicitly guarantee the feasibility of ground contact constraints. 
\end{abstract}



\section{Introduction}

Legged robots, mimicking nature's design of terrestrial animals, have many potential advantages compared to wheeled or tracked robots in traversing unstructured terrains.
Various control approaches have been created to enable stable and robust locomotion on stationary surfaces \cite{anymal_16,cheetah3}.
Yet, sustaining legged locomotion on a nonstationary surface has not been fully addressed.

The objective of this study is to derive a control approach that stabilizes locomotion on a dynamic rigid surface (DRS) by explicitly addressing the time-varying surface movement and the feasibility of ground contact forces.
A key element of the proposed approach is an online footstep planner that considers the surface motion through the extension of the concept of capture point \cite{pratt2006capture,Pratkoolen2012capturability_theory}
from stationary to dynamic surfaces.
Another key element is a quadratic programming (QP) based control method that incorporates the surface motion.
Related work is reviewed next.

\subsection{Related work on Legged Locomotion Control for Rigid Stationary and Deformable Surfaces}

Previous control approaches for legged robot locomotion mainly focus on static surfaces (e.g., pavement and stairs)~\cite{Kajita2003BipedWP,gu2018exponential,gu2018straight,gao2019global,yuan2019DSCC,barasuol2013reactive,anymal_16,cheetah3}.
Yet, they may not work well for locomotion on a deformable surface or a DRS because they do not explicitly account for the perturbations caused by the surface deformation or movement.

One of the pioneering efforts in addressing locomotion on deformable surfaces is the modeling of the dynamic interaction between the robot's legs and granular terrains (e.g., sand and snow)~\cite{terra_li2013ATO}.
Based on this model, Xiong et al.~\cite{Xiong2017ASR} have developed a stability region criterion for granular terrain walking to guide controller design.
However, it is unclear how effective these methods would be for sustaining locomotion on a DRS.
\subsection{Related Work on DRS Locomotion Control}
Stabilizing legged locomotion on a DRS is fundamentally challenging due to the high complexity of the associated robot dynamics.
First, the robot dynamics are inherently hybrid involving state-triggered discrete behaviors (e.g., foot touchdowns)~\cite{galeani2011robust}.
Second, if not properly addressed, the time-varying perturbation at a foot-surface contact point caused by the surface motion can destabilize the robot.

Recently, planning and control of DRS locomotion have been increasingly extensively studied.
For surfaces with a motion affected by the locomotor (e.g., passive surfaces with relatively small inertia),
planning and control methods for bipedal walking on different-sized balls have been proposed based on reduced-order~\cite{Yamenzheng2011ball} and full-order robot models~\cite{BallMan2020Koshil_yang}.
Controller design for legged locomotion on a floating island has also been studied based on an eight-legged rimless wheel robot model~\cite{asano2021modeling}.
Still, these methods cannot be directly used to handle locomotion on a DRS whose motion is not affected by the robot, due to the differences in the associated robot dynamics.

To address locomotion on a DRS whose motion is not affected by the robot (e.g., vessels and aircraft), our previous work~\cite{iqbal_SLIP} has analyzed the effects of surface movement on a robot's locomotion stability based on the reduced-order spring-loaded inverted pendulum (SLIP) robot model.
We have also recently derived and experimentally evaluated a provably stabilizing controller for enabling quadrupedal locomotion on a DRS~\cite{iqbal2020provably}.
Yet, this controller has two major limitations towards effective implementation in real-world scenarios.
First, the previous planning algorithm is not suitable for online motion generation due to the heavy computational load caused by its underlying full-order model.
Second, the controller does not explicitly account for the feasibility of ground-contact constraints during the actual walking process. 
\subsection{Related Work on Capture Point and QP-based Control}%
A capture point is defined as the point that an inverted pendulum model needs to instantaneously extend its leg to in order to come to a complete stop within the new step~\cite{pratt2006capture,Pratkoolen2012capturability_theory}.
Due to its computational efficiency,
the concept of capture point has been used to plan footsteps for legged locomotion on static~\cite{englsberger2011bipedal_CapPoint} and uneven terrains~\cite{Cap_Uneven_Morisawa2012balance,Cap_uneve_caron2019TRO} as well as for disturbance rejection~\cite{Cap_Ext_Pert_joe2018balance}.
However, to the best of our knowledge, the capture point has not been extended to solve the footstep planning problem for DRS locomotion.
Due to its capability of incorporating feasibility constraints and its low computational load suitable for real-time implementation, QP-based control design has been extensively studied for legged locomotion over high-slope terrains~\cite{QP_focchi2017high} and unstructured terrains~\cite{cheetah3}. 
Recently, a QP-based controller utilizing a full-order robot model has shown effectiveness in asymptotically tracking desired trajectories~\cite{RAL_2020quadrupedal_hamed}.
Still, these controllers mainly focus on static surface locomotion and thus do not explicitly consider the time-varying motion of a DRS.

\subsection{Contributions}

This paper aims to derive a planning and control approach that enables stable and robust quadrupedal walking on a DRS by extending our previous work~\cite{iqbal2020provably} and solving its two major limitations, which are:
a) lack of feasibility guarantees in the controller design and
b) heavy computational load of motion planning that prevents online implementation.

The main contributions of this paper are:
a) Extending the concept of capture point to DRS locomotion by explicitly considering the surface motion, and designing an online footstep planner based on the extended capture point. 
b) Formulating a QP-based controller that ensures the feasibility of contact forces and explicitly incorporates the surface motion in a time-varying holonomic constraint.
c) Validating the stability and robustness of the proposed approach through simulations of quadrupedal walking on a DRS with a rocking motion.
d) Providing comparative simulation results to demonstrate the improved robustness of the proposed method over our previous work. 

\section{HYBRID FULL-ORDER ROBOT MODEL}
\label{Section-Dyanmics}

This section presents the full-order model of a quadrupedal robot that walks on a DRS, which captures the dynamic behaviors of the robot's full degrees of freedom (DOFs).
The model serves as a basis for the proposed QP-based controller design.

A walking robot's dynamics are hybrid because walking naturally involves continuous-time motion (e.g., foot swinging) and discrete events (e.g., foot touchdowns).
During locomotion on a DRS, the robot's dynamics are also affected the time-varying movement of the surface.
The hybrid, time-varying robot dynamics are explained next.

\textbf{Continuous-phase dynamics.}
Let $\mathbf{q}$ be the robot's generalized coordinates, which comprise the base pose (i.e., position $\mathbf{p}_b$ and orientation $\boldsymbol{\gamma}_b$) and joint angles $\begin{bmatrix}
q_1~q_2~...~q_n
\end{bmatrix}^T$; that is,
$
    \mathbf{q}:=
    \begin{bmatrix}
    \mathbf{p}_b^T,~\boldsymbol{\gamma}_b^T,~q_1,~q_2,~...,~q_n \end{bmatrix} ^T.
$
Let $\mathcal{Q} \subset \mathbb{R}^{n+6}$ be the robot's configuration space. 
Then $\mathbf{q} \in \mathcal{Q}$.
Let $\mathbf{u}  \in U \subset \mathbb{R}^m$ be the robot's joint torques with $U$ the admissible set of joint torques.

Using Lagrange's method, the continuous-phase robot dynamics can be obtained as:
\begin{equation}
	\mathbf{M(q)\ddot{q}+C(q,\dot{q}) := Bu}+\mathbf{J}_f^T(\mathbf{q}) \mathbf{F}_f,
	\label{Eq:Rob_Dyn_ideal}
\end{equation}
where $\mathbf{M}(\mathbf{q}): \mathcal{Q} \rightarrow \mathbb{R}^{(n+6) \times (n+6)}$ is the inertia matrix,
$\mathbf{\mathbf{C}(\mathbf{q},\dot{\mathbf{q}})}:  \mathcal{TQ} \rightarrow \mathbb{R}^{(n+6)}$ represents the sum of centrifugal, Coriolis, and gravitational terms with $\mathcal{T Q} $ the tangent space of $\mathcal{Q}$,
and $\mathbf{B}\in \mathbb{R}^{(n+6) \times m}$ is the actuator selection matrix.
The matrix $\mathbf{J}_f (\mathbf{q})$ is the contact Jacobian whose expression varies during different phases because of different feet in support.
The vector
$\mathbf{F}_f (\mathbf{q}) \in \mathbb{R}^{n_{ct}}$ represents the external force acting at the support feet.
During three-dimensional ($3$-D) quadrupedal walking without slip, three of the four legs are in secured contact with the surface, and therefore $n_{ct}=9$.

During robot walking on a DRS whose motion is not affected by the robot (e.g., elevators and vessels),
the effects of the surface motion on the robot dynamics can be modeled as a time-varying  holonomic constraint.

Let {$\mathbf{r}_p(t) \in \mathbb{R}^{n_{ct}}$} be the positions of the surface at the support-foot locations. 
Note that for a DRS, the surface positions and motions at the foot-contact points (i.e., $\mathbf{r}_p(t)$ and its derivatives) are explicitly time-varying.
Let $\mathbf{r}_f(\mathbf{q}) \in \mathbb{R}^{n_{ct}}$ be the positions of the support feet.
The contact Jacobian matrix, by definition, is expressed as $\mathbf{J}_f (\mathbf{q}) :=
\frac{\partial \mathbf{r}_f}{\partial \mathbf{q}}  (\mathbf{q})$.

When the support feet do not slip on the surface, the time-varying holonomic constraint can be formed as:
\begin{equation} 
    \mathbf{J}_f (\mathbf{q}) \ddot{\mathbf{q}} + \mathbf{\dot{J}}_f ((\mathbf{q}),\dot{\mathbf{q}}) \dot{\mathbf{q}}
    = \ddot{\mathbf{r}}_p(t).
  	\label{Eq:DD-DP_holonomic_const}
\end{equation}

\textbf{Switching surface}
When a swing foot strikes the walking surface, it becomes a new stance foot while another foot begins to swing in the air and acts as the new swing foot.
The foot landing event can be defined through the switching surface,
$
    S:=\{ (\mathbf{q},\dot{\mathbf{q}},\mathbf{r}_p,\dot{\mathbf{r}}_p) \in \mathcal{T Q} \times \mathbb{R}^{n_{ct}} \times \mathbb{R}^{n_{ct}}:
    \phi(\mathbf{q},\mathbf{r}_p)=0,
    ~\dot{\phi}(\mathbf{q},\dot{\mathbf{q}},\mathbf{r}_p,\dot{\mathbf{r}}_p)<0\},
$
where the scalar function $\phi$ is the distance  between the swing foot and the surface.

\section{Capture-point based Online Planning}
\label{Section-Planning}

This section introduces the extension of the capture point from stationary to DRS walking, as well as the proposed online planning method based on the extended capture point.
The desired trajectories produced by the planner will be tracked by the proposed QP-based control as introduced in Sec.~\ref{Section-Control}.

\subsection{Capture Point on a DRS}
\label{SUBSection-ONLINEPlanning}

\begin{figure}[t]
    \centering
    \includegraphics[width= 0.82\linewidth]{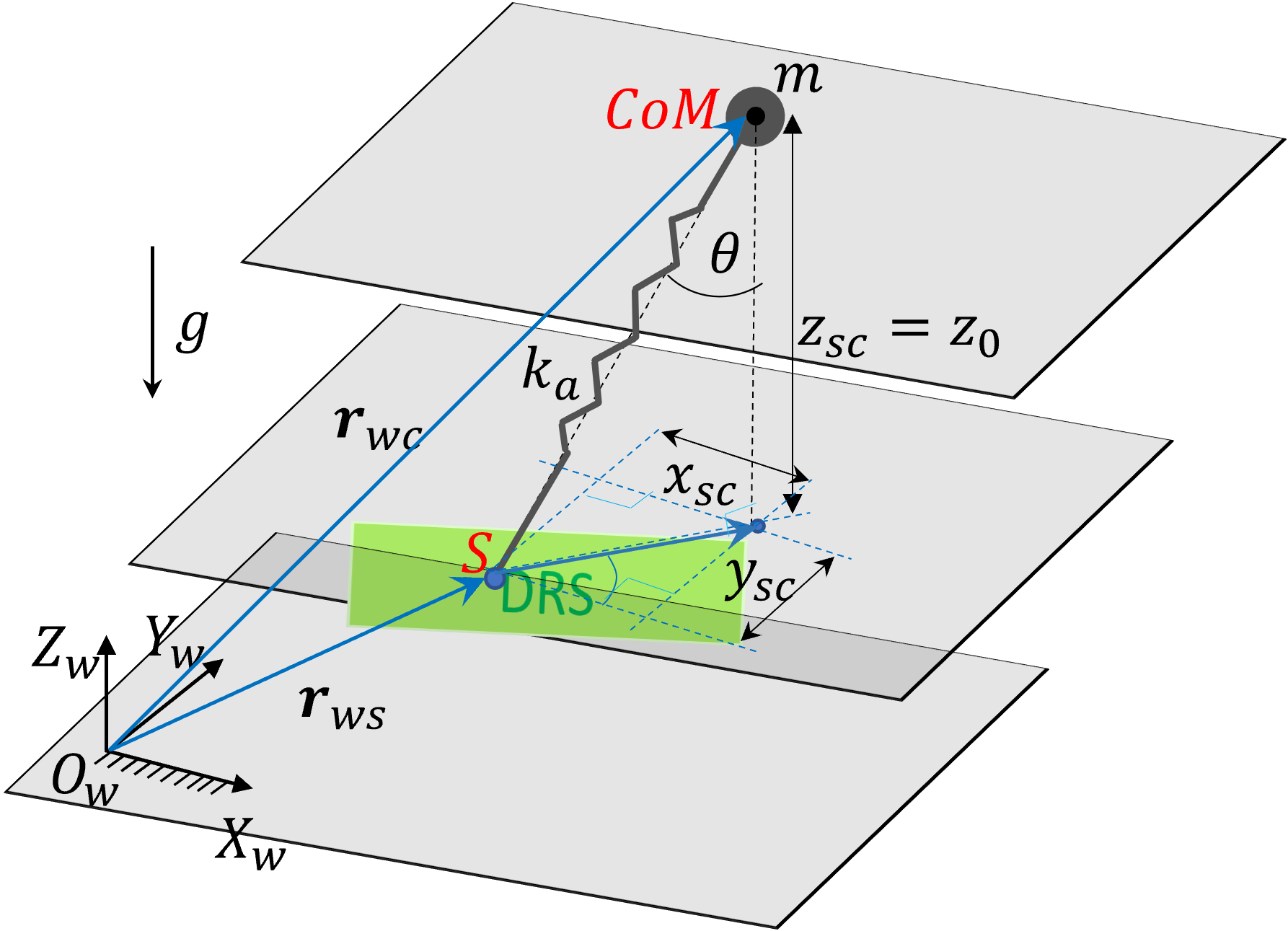}
    \caption{Illustration of the 3-D LIPM model on a DRS. The model is constrained to maintain constant height $z_0$ between the CoM and the point S on the DRS.
    }
    \label{Fig:LIPM_sketch}
\end{figure}

The concept of capture point for a static surface has been extensively studied \cite{pratt2006capture,Pratkoolen2012capturability_theory}.
In this work, we extend the capture point to a DRS based on a reduced-order locomotor model, that is, a $2$-D linear inverted pendulum model (LIPM) with a point mass atop a massless leg (see Fig.~\ref{Fig:LIPM_sketch}).
The position of the point mass coincides with the robot's center of mass (CoM).

Let $\mathbf{r}_{wc}=(x_{wc},~x_{wc},~z_{wc})$ and {$\mathbf{r}_{ws}=(x_{ws},~x_{ws},~z_{ws})$} be the absolute positions of the CoM and the leg's far end (i.e., point $S$) in the world frame, respectively.
Then, the CoM position relative to point $S$ is given by: 
\begin{equation}
    \mathbf{r}_{sc}=(x_{sc},~x_{sc},~z_{sc})
    = \mathbf{r}_{wc} -\mathbf{r}_{ws}.
    \label{Eq:CoordinateTransf}
\end{equation}

The equation of motion of the LIPM is expressed as:
\begin{align}
    \ddot{x}_{wc} = \tfrac{f_a x_{sc}}{m r} &\sin \theta,
    ~ \ddot{y}_{wc} = \frac{f_a y_{sc}}{m r} \sin \theta ,
    ~
    \ddot{z}_{wc} = \frac{f_a}{m} \cos \theta -g, 
     \label{Eq: LIPM_EOM_world - z}
\end{align}
where $m$ is the mass of the LIPM, $\theta$ is the angle of the leg relative to the vertical axis, $g$ is the magnitude of the gravitational acceleration, and $r$ is the projected length of $\mathbf{r}_{sc}$ on the horizontal plane.
The scalar variable $f_a$ is the norm of the axial force that the massless rod applies to the point mass. 

It is assumed in this study that the vertical distance $z_{sc}$ between the CoM and point $S$ is constant during walking (see Fig.~\ref{Fig:LIPM_sketch});
that is, $z_{sc} = z_0$ with $z_0$ a positive number.
This assumption is analogous to the simplifying assumption of the LIPM model on a stationary surface that the point-mass height over the surface is constant~\cite{pratt2006capture}.

Under the assumption that $z_{sc}$ is constant, we have
$\dot{z}_{wc} = \dot{z}_{ws}$ and $\ddot{z}_{wc} = \ddot{z}_{ws}$.
Then, from \eqref{Eq: LIPM_EOM_world - z}, the expression of $f_a$ becomes
$
     f_a = m \frac{(\ddot{z}_{ws}+g)}{\cos \theta}.
     \label{Eq:axialForce}
$
Accordingly, the equation of motion of the LIPM on the horizontal plane is now given by:
    \begin{align}
     \ddot{x}_{wc} =
     (\ddot{z}_{ws}+g)\frac{x_{sc}}{z_0}
     ~\text{and}~
    \ddot{y}_{wc}= 
    (\ddot{z}_{ws}+g)\frac{y_{sc}}{z_0}.
     \label{Eq:XY-motion general}
    \end{align}
From~\eqref{Eq:CoordinateTransf}, we obtain $\ddot{x}_{wc}=\ddot{x}_{ws}+ \ddot{x}_{sc}$ and $\ddot{y}_{wc}=\ddot{y}_{ws}+ \ddot{y}_{sc}$.
Then, using these relations in \eqref{Eq:XY-motion general} yields:
\begin{align}
    \ddot{x}_{sc}  - \frac{(\ddot{z}_{ws}+g)}{z_0} x_{sc} = \ddot{x}_{ws}
   ~ \text{and}~
    \ddot{y}_{sc}  - \frac{(\ddot{z}_{ws}+g)}{z_0} y_{sc} = \ddot{y}_{ws}.
    \label{Eq:simplified_Cap-xy}
\end{align}

The ordinary differential equations in \eqref{Eq:simplified_Cap-xy} are non-homogeneous.
If the horizontal accelerations of point $S$ (i.e., $\ddot{x}_{ws}$ and $\ddot{y}_{ws}$) are sufficiently small,
the forcing terms can be approximated as zero, and these equations become homogeneous:
\begin{align}
    \ddot{x}_{sc}  - \frac{(\ddot{z}_{ws}+g)}{z_0} x_{sc} = 0
    \label{Eq:simplified_Cap-2_x} ~\text{and}~
    \ddot{y}_{sc}  - \frac{(\ddot{z}_{ws}+g)}{z_0} y_{sc} = 0.
\end{align}

Analogous to the derivation of the capture point on a stationary surface {\cite{pratt2006capture}}, we can define the apparent stiffness of the LIPM as $k_a:=-\frac{(\ddot{z}_{ws}+g)}{z_0}$.
Then, we reduce the LIPM's equation to a spring-mass system with unit mass; that is, $ \ddot{x}_{sc}  + k_a x_{sc}=0$ and $ \ddot{y}_{sc}  + k_a y_{sc}=0$.

The orbital energy of this system associated with the horizontal motion is:
$$
    E_x =\frac{1}{2}(\dot{x}_{sc} ^{2} + k_a x_{sc}^{2}) ~\text{and}~
    E_y =\frac{1}{2}(\dot{y}_{sc} ^{2} + k_a y_{sc}^{2}).
$$
These energy terms predict whether the point mass will move over the support point $S$ (i.e., $E_x+E_y>0$), stop at $S$ (i.e., $E_x+E_y=0$), or reverse the direction of motion before reaching $S$ (i.e., $E_x+E_y<0$).

At the equilibrium (i.e., $E_x=0$ and $E_y=0$) of the system in~\eqref{Eq:simplified_Cap-2_x},
the two eigenvectors are:
\begin{align}
    \dot{x}_{sc} = \pm \sqrt{\frac{(\ddot{z}_{ws}+g)}{z_0}}x_{sc} ~\text{and}~
    \label{Eq:Egenvectors_x} 
    \dot{y}_{sc} = \pm \sqrt{\frac{(\ddot{z}_{ws}+g)}{z_0}}y_{sc}.
\end{align}

The stable eigenvector corresponds to the case when $x_{sc}$ and $\dot{x}_{sc}$ (and $y_{sc}$ and $\dot{y}_{sc}$) have opposite signs, which lead to the following horizontal coordinates of the instantaneous capture point 
of the $3$-D LIPM on a DRS:
\begin{align}
    x_{cap} :=  \sqrt{\frac{z_0}{(\ddot{z}_{ws}+g)}}\dot{x}_{sc} ~\text{and}~
    \label{Eq: CapturePoint_DP-x}
    y_{cap} :=  \sqrt{\frac{z_0}{(\ddot{z}_{ws}+g)}}\dot{y}_{sc}.
\end{align}
The vertical coordinate of the instantaneous capture point lies on the DRS.
For a horizontal, flat DRS with a constant horizontal velocity but a varying vertical velocity (e.g., elevators), the vertical coordinate of the capture point will be zero, i.e., $z_{cap} =0$.

The capture point $(x_{cap},y_{cap},z_{cap})$ is the position of the new support foot relative to the previous one. 
Note that the capture point no longer exists when the term $(\ddot{z}_{wp}+g)$ is negative (i.e., the robot loses contact with the DRS).

Analogous to the capture point on a stationary surface, if the LIPM instantaneously places the far end of its leg (point $S$) at $(x_{cap},y_{cap},z_{cap})$, it will come to a complete stop {\it relative to the new support point} within the immediate successive step.


{The expression of the horizontal coordinates of the capture point in
\eqref{Eq: CapturePoint_DP-x} 
explicitly considers the surface motion that was not incorporated in the previous derivations of the capture point on a stationary surface~\cite{pratt2006capture,Cap_uneve_caron2019TRO}.}
Note that when $\ddot{z}_{ws} = 0$ (i.e., the surface is stationary or moving with a constant speed in both the horizontal and vertical directions), the proposed capture point on a DRS in \eqref{Eq: CapturePoint_DP-x} 
reduces exactly to the capture point on a stationary surface.

\subsection{Capture-Point based Online Planning}

The objective of the proposed footstep planner is to generate a robot's desired foot placement for maintaining a certain desired walking motion {\it instead of} forcing the robot to stop.
To this end, the capture point $(x_{cap},y_{cap},z_{cap})$ is further extended to determine the desired footsteps that help sustain the planned robot movement.
For simplicity, the desired motion is specified by a constant base velocity $\mathbf{v}_d = (v_{dx},v_{dy},0)$ relative to point $S$.

We choose to design the desired horizontal foot placement ($\tilde{x}_{step},\tilde{y}_{step}$)
as the difference between the capture point associated with the current relative CoM velocity $\mathbf{\dot{r}}_{sc}$ and a fictitious capture point corresponding to the desired velocity $\mathbf{v}_d$~\cite{cheetah3} multiplied by a positive gain $K_{step}$:
\begin{equation}
\textcolor{black}{
\begin{aligned}
        \tilde{x}_{step}
        &:=
        K_{step}\sqrt{\frac{z_0}{(\ddot{z}_{ws}+g)}}(\dot{x}_{sc}-v_{dx})
        ~\text{and}~
        \\
\tilde{y}_{step}
        &:=
        K_{step} \sqrt{\frac{z_0}{(\ddot{z}_{ws}+g)}}(\dot{y}_{sc}-v_{dy}).
        \label{Eq:Cap_criterion}
\end{aligned}
}
\end{equation}
The gain $K_{step}$ can be used to tune how much the difference between the current and desired CoM velocities affects the desired footstep location, thus allowing us to consider practical hardware limitations (e.g., finite step duration and limited leg length) in footstep planning.

To ensure that the robot moves with $v_{dx}$ and $v_{dy}$ in the horizontal directions, Raibert's heuristic {\cite{cheetah3}}
is used along with the instantaneous capture point in~\eqref{Eq:Cap_criterion} to plan the desired foot placement:
\begin{equation}
    {x}_{step} = {x}_{hip} + \frac{T}{2}v_{dx} +
    \tilde{x}_{step},
    ~
    {y}_{step} = {y}_{hip} + \frac{T}{2}v_{dy} + 
    \tilde{y}_{step},
    \label{Eq: swingfoot_planning_cap point-x}
\end{equation}
where $(x_{hip},y_{hip})$ is the horizontal coordinates of the swing hip joint and $T$ is the desired gait period.

The $z$-coordinate of the desired foot placement on a horizontal, flat DRS is ${{z}}_{step} = \tilde{z}_{step} = 0$.

Given the desired CoM trajectory and foot placement, we utilize inverse kinematics to generate the desired full-body trajectories.
Since whole-body trajectory generation is not the focus of this study, its details are omitted for brevity.

\section{Instantaneous QP-based Controller Design}
\label{Section-Control}
This section introduces the proposed QP-based controller design that provably tracks the desired walking motions $\mathbf{h}_d(t)$ on a DRS and guarantees the feasibility for ground contact forces.
The QP-based controller is synthesized based on full-order dynamics of quadrupedal walking on a DRS, and explicitly addresses the surface movement through the formulation of the holonomic and the friction cone constraints.

Both QP and MPC can be used to formulate a controller with feasibility guarantees. 
We choose to use QP because its computational load is low enough for the typical real-time controller implementation rate (e.g., $500$ Hz or higher) even when the QP is formulated based upon a full-order robot model.
In contrast, an MPC {synthesized based on a full-order model} typically computes its output at a much lower rate, e.g., $30$Hz.

The key steps to the proposed QP formulation include the selection of optimization variables and the design of the cost function and constraints, which are explained next.


\textbf{Optimization variables.}
As the control objective is to achieve reliable trajectory tracking with the ground contact constraints respected,
we define the optimization variables $\mathbf{x}$ as the control input $\mathbf{u}$ and ground contact force $\mathbf{F}_f$, i.e., $\mathbf{x}=[\mathbf{u}^T,\mathbf{F}_c^T]^T$.

\textbf{Cost function.}
The {quadratic} cost function to be minimized is set as
$ \mathbf{x}^T\mathbf{Q}\mathbf{x} + (\tilde{\mathbf{x}}-\mathbf{x})^T \mathbf{W}(\tilde{\mathbf{x}}-\mathbf{x})$ with $\tilde{\mathbf{x}}$ the values of the optimization variables at the previous {optimization} step.
The weighting matrix $\mathbf{Q}$ is a symmetric, positive definite matrix with an appropriate dimension. 
The minimization of $\mathbf{x}^T\mathbf{Q}\mathbf{x}$ is to reduce the demanded joint torques. 
The other weighting matrix, $\mathbf{W}$, is a symmetric, positive definite matrix that can be used to penalize sharp variations in some of the optimization variables.

\noindent \textbf{Equality time-varying holonomic constraints:}
To ensure that the support feet do not move relative to the surface,
we consider the time-varying holonomic contraint in~\eqref{Eq:DD-DP_holonomic_const} that explicitly contains the surface movement.

From the continuous-phase dynamics in \eqref{Eq:Rob_Dyn_ideal}, we have
   $ \label{eq: ddq}
    \mathbf{\ddot{q}} := \mathbf{M}^{-1}(\mathbf{B u}+\mathbf{J}_f^T \mathbf{F}_f-\mathbf{C}).$
Substituting it in the holonomic constraint in \eqref{Eq:DD-DP_holonomic_const} gives:
$
    \mathbf{J}_f \mathbf{M}^{-1}(\mathbf{Bu}+\mathbf{J}_f^T  \mathbf{F}_f-\mathbf{C}) + \mathbf{\dot{J}}_f  \dot{\mathbf{q}}
    = \ddot{\mathbf{r}}_p(t),
  	\label{Eq:DE_holonomic_const}
$
which, upon rearrangement, gives:
    $\mathbf{A}_{eq1}\mathbf{x} =\mathbf{b}_{eq1},
    \label{Eq:EC1_Holonomic}$
where $\mathbf{A}_{eq1} := [\mathbf{J}_f \mathbf{M}^{-1} \mathbf{B},~ \mathbf{J}_f \mathbf{M}^{-1} \mathbf{J}_f^T]$
and $\mathbf{b}_{eq1} := \mathbf{J}_f \mathbf{M}^{-1} \mathbf{C} - \mathbf{\dot{J}}_f \dot{\mathbf{q}} + \ddot{\mathbf{r}}_p(t)$.

\noindent \textbf{Equality constraints based on input-output linearizing control:} 
To realize accurate tracking of the desired base and swing-foot trajectories $\mathbf{h}_d(t)$, the controller takes the form of an input-output (I-O) linearizing control law~{\cite{iqbal2020provably}} with the output function defined as the tracking error.

Let $\mathbf{h}$ denote the variables of interest (i.e., base pose and swing foot position).
Then, the output function representing the tracking error is given by: 
    $\mathbf{y} := \mathbf{h}(\mathbf{q}) - \mathbf{h}_d(t).
    \label{Eq:OutputFunction}$
Then, 
    $\mathbf{\dot{y}} 
    = 
    \frac{\partial{\mathbf{h}}}
    {\partial \mathbf{q}} \dot{\mathbf{q}} 
    - \dot{\mathbf{h}_d} 
	\label{derivative_OF}
$
and 
$\mathbf{\ddot{y}} 
    = \frac{\partial}{\partial \mathbf{q}}
    (\frac{\partial{\textbf{h}}}{\partial \mathbf{q}}\dot{\mathbf{q}})
    \dot{\mathbf{q}} 
    + 
    \frac{\partial{\mathbf{h}}}
    {\partial \mathbf{q}} \ddot{\mathbf{q}} 
    - \ddot{\mathbf{h}}_d 
	\label{dderivative_OF}$.
Define $\mathbf{J}_{h}:=\frac{\partial{\mathbf{h}}}
{\partial \mathbf{q}}$, and then $\mathbf{\dot{J}}_{h}=\frac{\partial}{\partial \mathbf{q}}
(\frac{\partial{\textbf{h}}}{\partial \mathbf{q}}\dot{\mathbf{q}})$.
We get the output function dynamics as $\mathbf{\ddot{y}} = \mathbf{J}_{h} \ddot{\mathbf{q}} + \mathbf{\dot{J}}_{h}\dot{\mathbf{q}}-\ddot{\mathbf{h}}_d$.

To stabilize the output function dynamics, we utilize a proportional derivative (PD) feedback term
$\label{eq: v}
        \mathbf{\ddot{y}} = -\mathbf{K}_p \mathbf{y} -\mathbf{K}_d\mathbf{\dot{y}} =: \mathbf{v},$
which, combined with the expression of $\ddot{\mathbf{q}}$ in \eqref{eq: ddq}, yields:
        $\mathbf{J}_{h} \mathbf{M}^{-1}(\mathbf{Bu}+\mathbf{J}_c^T\mathbf{F}_c-\mathbf{C}) + \mathbf{\dot{J}}_{h}\dot{\mathbf{q}}+\ddot{\mathbf{h}}_d (t) = \mathbf{v}.
        \label{Eq:virtual_Const_Dyn}$
It can be compactly written as
    $\mathbf{A}_{eq2}\mathbf{x} =\mathbf{b}_{eq2},
    \label{Eq:EC2_Virtual_CD}$
where $\mathbf{A}_{eq2} := [\mathbf{J}_{hc} \mathbf{M}^{-1} \mathbf{B},~ \mathbf{J}_{h} \mathbf{M}^{-1} \mathbf{J}_f^T]$
and
$\mathbf{b}_{eq2} := \mathbf{J}_{hc} \mathbf{M}^{-1} \mathbf{C} -\mathbf{\dot{J}}_{h}\dot{\mathbf{q}}-\ddot{\mathbf{h}}_d(t) + \mathbf{v}$.

\noindent \textbf{Inequality ground-contact constraints {and joint torque limits.}}
To secure contacts between support feet and the surface, the QP incorporates the friction cone (i.e., no foot slipping) and unilateral (i.e., no feet penetrating the surface) constraints
$
    \mathbf{F}_f \in \mathcal{F}_{gc},
    \label{Eq:FrictionCone_Unilateral}
$
where $\mathcal{F}_{gc}$ is the set of ground-contact forces satisfying the friction cone and unilateral constraints.
Also, the solution to the QP must respect the torque limits; i.e., the control input $\mathbf{u}$ must be within the admissible torque set, i.e., $\mathbf{u} \in U$.

\noindent \textbf{QP-based control law.}
With the cost function and constraints designed, the proposed QP 
can be compactly expressed as:
\begin{equation}
\begin{aligned}
\min_{\mathbf{x}} \quad & \mathbf{x}^T\mathbf{Q}\mathbf{x} + (\mathbf{x}_p-\mathbf{x})^T \mathbf{W}(\mathbf{x}_p-\mathbf{x})\\
\textrm{subject to} \quad &  \mathbf{A}_{eq1}\mathbf{x} =\mathbf{b}_{eq1},
~\mathbf{A}_{eq2}\mathbf{x} =\mathbf{b}_{eq2},
~
  \mathbf{F}_f \in \mathcal{F}_{gc},~
  u \in \mathbf{U}.   
  \label{Optimization_U}
\end{aligned}
\end{equation}


\noindent \textbf{Stability property.} 
As proven in our previous study~\cite{iqbal2020provably},
an input-output linearizing controller with the proposed form can guarantee the (local) asymptotic stability of the desired trajectory $\mathbf{h}_d(t)$ for the hybrid, time-varying walking process, if the PD gains are chosen to render a sufficiently fast error convergence rate during continuous phases.

\section{Simulations}
This section presents MATLAB simulation results that validate the effectiveness of the proposed control approach.

\noindent\textbf{Robot.}
{The simulated robot is a Laikago quadruped developed by Unitree Robotics. 
Its total mass is 25 kg, and each leg weighs 2.9 kg.
Each leg has three independently actuated joints (i.e., hip-roll, hip-pitch, and knee-pitch) with torque limits of 20 Nm, 55 Nm, and 55 Nm, respectively.
The length, width, and height of the robot are 0.56m, 0.35m, and 0.60m, respectively.}
\begin{figure}[t]
    \centering
    \includegraphics[width= 0.8\linewidth]{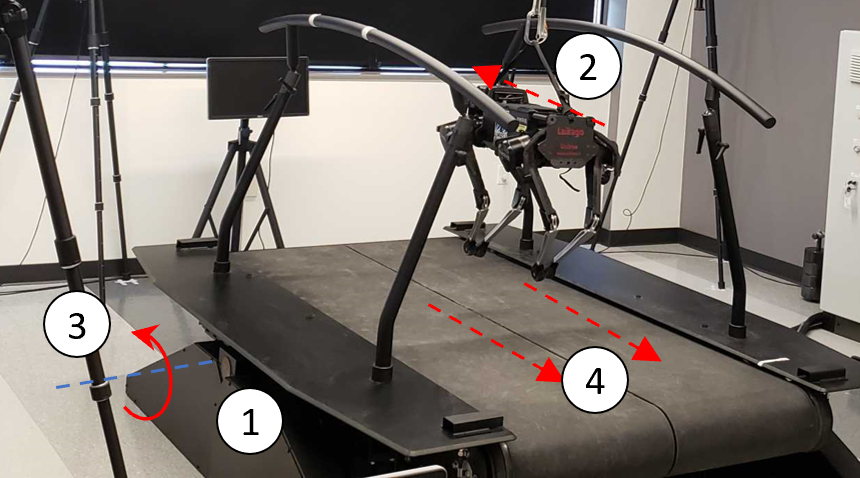}
    \caption{System setup that the simulator emulates.
    The treadmill (\textcircled{1}) has a split belt (\textcircled{4}) that moves at a constant speed of $8$cm/s while the treadmill undergoes a sinusoidal rocking motion about the horizontal axis (\textcircled{3}).
    The quadruped is a Laikago robot (\textcircled{2}) developed by Unitree Robotics.
    }
    \label{Fig:Simulated_exp_setup}
\end{figure}

\noindent \textbf{Surface.}
The simulated DRS is an actuated platform (see Fig.~\ref{Fig:Simulated_exp_setup}) simultaneously experiencing: a) a whole-body sinusoidal pitching motion, with an amplitude of $\pm 5^{\circ}$ and frequency of $0.5$ Hz and b) a constant surface translating motion.
This DRS reasonably satisfies the assumption underlying the proposed capture point extension, because the horizontal velocity of the surface is approximately constant due to the small pitching amplitude.
The surface acceleration in the vertical direction is still relatively significant for validating the proposed method. 

\noindent \textbf{QP-based controller.}
{The PD gains are $\mathbf{K}_p= 120 \cdot \mathbf{I}_9$ and $\mathbf{K}_d= 22 \cdot \mathbf{I}_9$, where $\mathbf{I}_n$ is an $n \times n$ identity matrix.
The weighting matrices in the cost function are set as: $\mathbf{Q}=\text{blockdiag}(1000 \cdot \mathbf{I}_{12},~\mathbf{I}_9,)$ and
$\mathbf{W}=\text{blockdiag}(10^{-1} \cdot \mathbf{I}_{12},~10^{-4} \cdot \mathbf{I}_9,)$.
The torque bounds are set based on the robot's hardware limits.
In the unilateral constraint, the lower bound of the normal ground reaction force at a support foot is set as $1$ N.}
In the friction cone constraint, the friction coefficient is chosen as 0.5.

\noindent \textbf{Planner.}
The inputs to the planner are:
a) gait cycle $T=2$ s, 
b) maximum swing foot height $15$ cm,
c) walking speed $8$ cm/s,
and d) the surface movement as explained earlier.
The gain used to tune the desired foot placement is set as $K_{step}=1$.

\begin{figure}[t]
    \centering
    \includegraphics[width=0.99\linewidth]{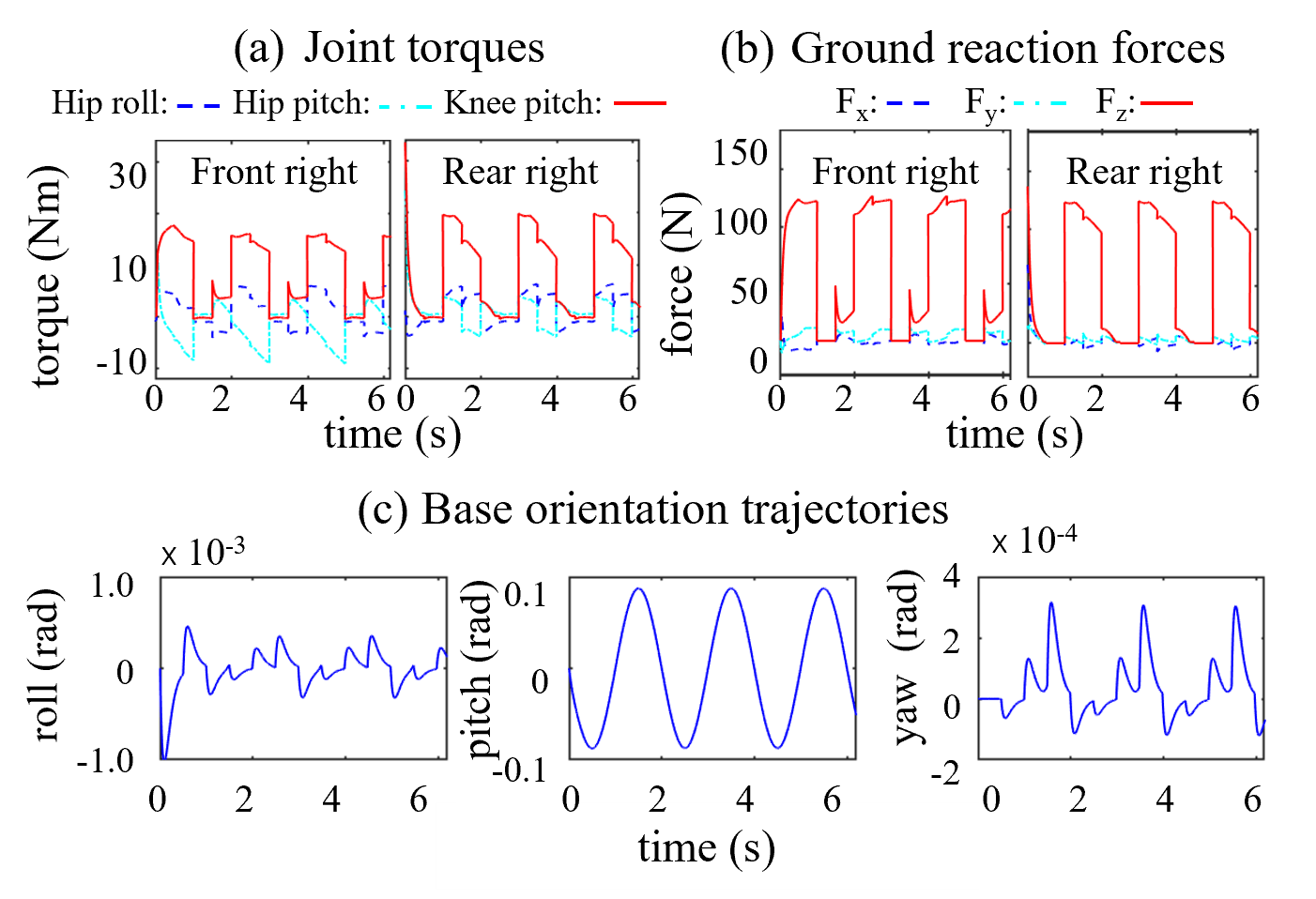}
    \caption{Simulation results obtained from unperturbed walking (Case 1): (a) joint torques, (b) ground reaction forces, and (c) base trajectories.
    }
    \label{Fig:Unperturbed}
\end{figure}

\begin{figure}[t]
    \centering
    \includegraphics[width=0.99\linewidth]{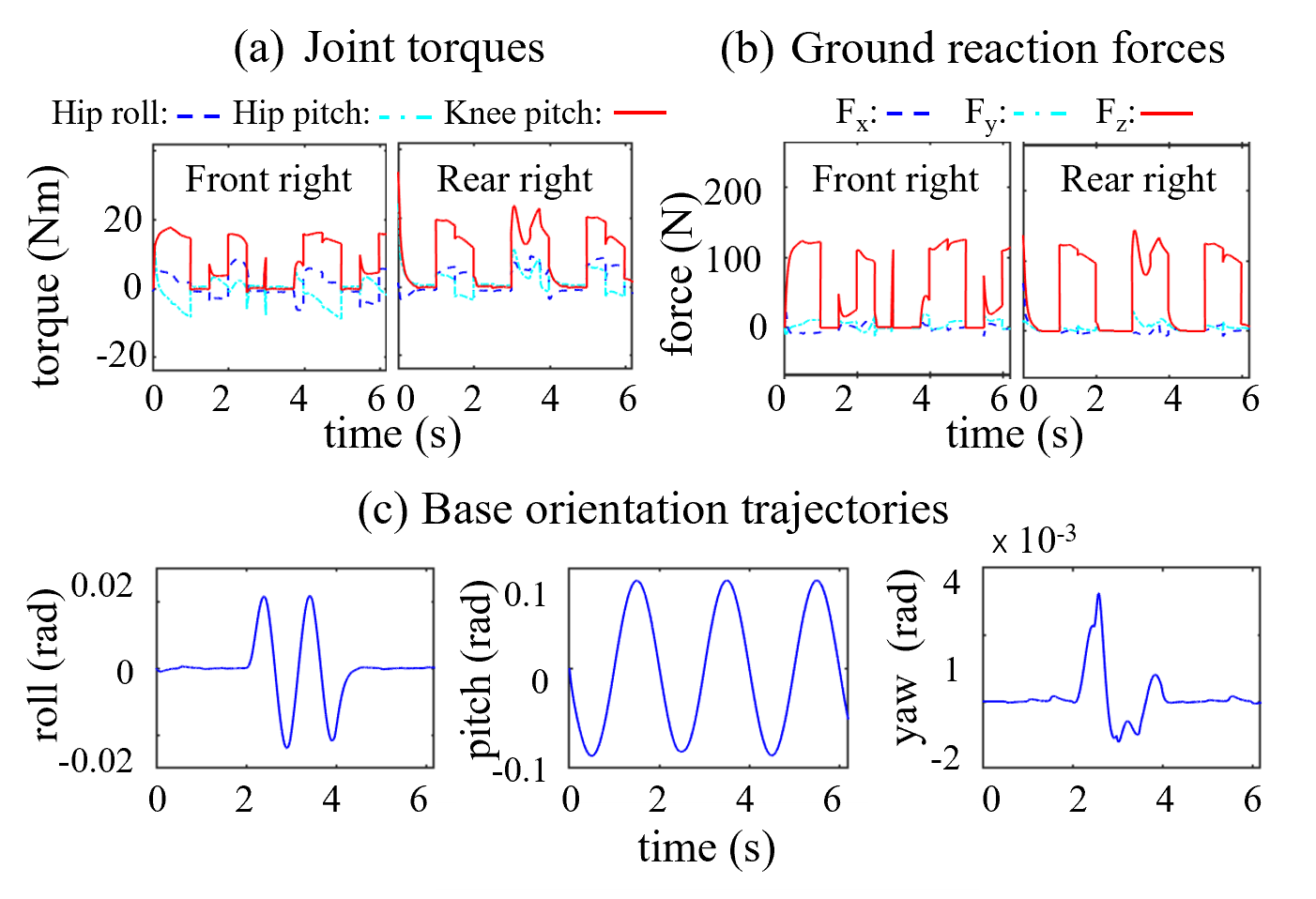}
    \caption{Simulation results obtained in the presence of lateral perturbations (Case 2):
    (a) joint torques, (b) ground reaction forces, and (c) base trajectories. 
    }
    \label{Fig:Perturbation-Y}
\end{figure}

\begin{figure}[h]
    \centering
    \includegraphics[width=0.99\linewidth]{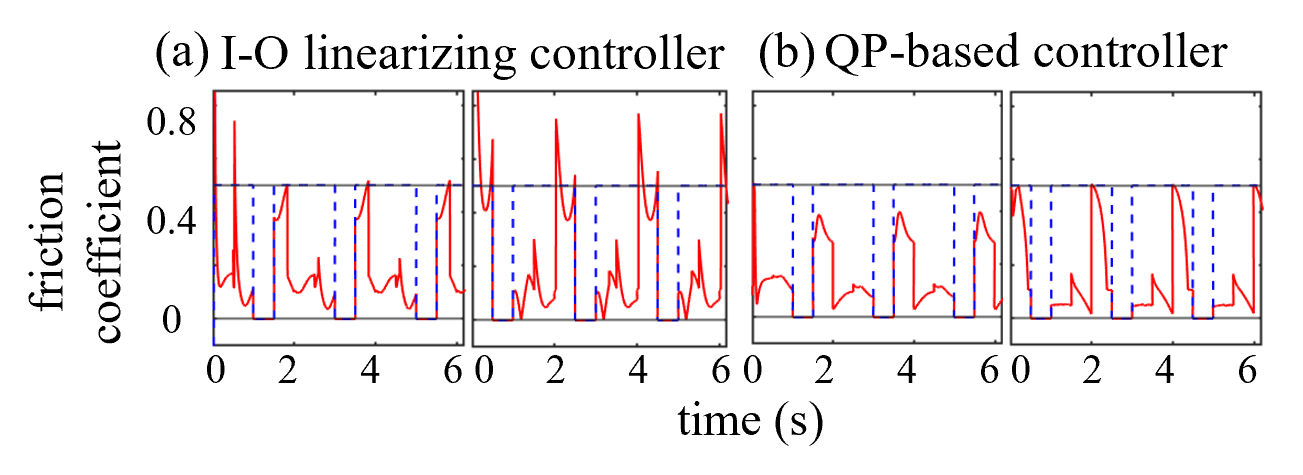}
    \caption{
    Comparative simulation results on the satisfaction of the friction cone constraint at the front right and rear right feet: (a) our previous offline planning and I-O linearizing control~\cite{iqbal2020provably} and
    (b) the proposed online planning and QP-based control.
    }
    \label{Fig:Comparative_result}
\end{figure}
\textbf{Case 1 (Unperturbed walking).}
Figure \ref{Fig:Unperturbed} shows the simulation results of DRS walking without perturbations under the proposed method. 
As shown in Fig.~\ref{Fig:Unperturbed} (c), 
the base orientation mildly varies, indicating that the proposed method can sustain stable walking on the DRS in the absence of perturbations.
Also, Figs.~\ref{Fig:Unperturbed} (a) and (b) show that the joint torques are within the limits and that the ground contact constraints are satisfied.

\textbf{Case 2 (Perturbed walking).}
Figure~\ref{Fig:Perturbation-Y} displays the simulation results of robot walking on the DRS under sinusoidal perturbation forces applied at the center of the robot's base. The magnitude of the external force is:
\begin{equation}
F_{p}(t) =
    \begin{cases}
    30\sin(2\pi t)~\text{N} &\text{if $2\text{s}\leq t \leq 4\text{s}$} \\
    0~\text{N} &\text{otherwise}
    \end{cases}
    \label{Eq:Perturb_force}
\end{equation}
The robot's base trajectories are notably perturbed between 2 sec and 4 sec, as shown in subplot (c).
In particular, the lateral base position trajectory is significantly displaced with a peak deviation of 0.12 m. 
Yet, the proposed control approach is able to sustain stable walking during the perturbations, as well as to drive the actual based trajectories to converge back to the desired motion after the perturbation is over. 
Also, the joint torque and ground contact constraints are respected throughout the walking process, as demonstrated in subplots (a) and (b).

\textbf{Case 3 (Comparative results).}
Figure~\ref{Fig:Comparative_result} shows the simulation results of our previous offline planning and I-O linearizing control~\cite{iqbal2020provably} under unperturbed walking.
Although this controller explicitly addresses the surface motion, it relies on offline planning and does not explicitly guarantee the feasibility of ground contact forces.
The unilateral ground contact constraint is indeed respected by the controller during simulated walking, as shown in Fig.~\ref{Fig:Comparative_result} (a).
Yet, the friction cone constraint is violated at the front right and rear right feet.
In contrast, the proposed online planning and QP-based control approach is able to respect all ground contact constraints during the unperturbed walking (subplot (a) of Fig.~\ref{Fig:Comparative_result}), as well as during perturbations (subplots (b) in Fig.~\ref{Fig:Perturbation-Y}).

\section{Conclusion}
This paper has introduced an online planner and QP-based controller design that sustains stable quadrupedal walking on a DRS (e.g., elevators and ships) by explicitly considering the surface motion and guaranteeing the feasibility of ground contact constraints. 
The key element of the online planner is a real-time footstep generator synthesized based on the extension of the concept of capture point from stationary surfaces to a DRS.
The capture point was derived based on a 3-D LIPM model walking on a DRS that undergoes a constant horizontal velocity and a varying vertical motion.
The controller was formulated as a QP based on the full-order robot model that captures the support foot movement caused by the surface motion.
The effectiveness of the proposed approach in walking stabilization and disturbance rejection was demonstrated via simulations of 3-D quadrupedal walking over a DRS.
The future work is dedicated to experimental validation of the proposed approach on a physical quadrupedal robot walking on a real-world nonstationary surface with an uncertain motion. 

	
\balance
\bibliography{amir21a}

\bibliographystyle{ieeetr}
	
\end{document}